%% file: main.tex
\documentclass[10pt,twocolumn,letterpaper]{article}

\usepackage{iccv}              


\definecolor{iccvblue}{rgb}{0.21,0.49,0.74}
\usepackage[pagebackref,breaklinks,colorlinks,allcolors=iccvblue]{hyperref}

\usepackage{multirow}
\usepackage{colortbl}  
\usepackage{xcolor}
\usepackage{array}   

\usepackage[most]{tcolorbox}
\usepackage[x11names]{xcolor}  
\definecolor{lightbeige}{HTML}{F5F5DC}  

\usepackage{amssymb}
\usepackage{fontawesome}  

\usepackage{cuted}

\def\paperID{*****} 
\def\confName{ICCV}
\def\confYear{2025}

\title{Can MLLMs Read the Room? A Multimodal Benchmark for Verifying Truthfulness in Multi-Party Social Interactions}

\author{%
    Caixin Kang$^{1}$, Yifei Huang$^{1}$, Liangyang Ouyang$^{1}$, Mingfang Zhang$^{1}$, Yoichi Sato$^{1}$ \\
  $^{1}$ The University of Tokyo\\
  \texttt{\{cxkang,hyf,oyly,mfzhang,ysato\}@iis.u-tokyo.ac.jp}\\
}

\begin{document}
\maketitle
\input{sec/0_abstract}    
\input{sec/1_intro}
\input{sec/2_Related_Work}
\input{sec/3_Task_Dataset}

\input{sec/4_Benchmark}

\input{sec/5_Discussion_and_Conclusion}
{
    \small
    \bibliographystyle{ieeenat_fullname}
    \bibliography{main}
}


\end{document}

%% file: sec/0_abstract.tex
\begin{abstract}
As AI systems become increasingly integrated into human lives, endowing them with robust social intelligence 
has emerged as a critical frontier. A key aspect of this intelligence is discerning truth from deception, a ubiquitous element of human interaction that is conveyed through a complex interplay of verbal language and non-verbal visual cues. 
However, automatic deception detection in dynamic, multi-party conversations remains a significant challenge. 
The recent rise of powerful Multimodal Large Language Models (MLLMs), with their impressive abilities in visual and textual understanding, makes them natural candidates for this task. 
Consequently, their capabilities in this crucial domain are mostly unquantified. To address this gap, we introduce a new task, \textbf{Multimodal Interactive Veracity Assessment (MIVA)}, and present a novel multimodal dataset derived from the social deduction game \textbf{Werewolf}. This dataset provides synchronized video, text, with verifiable ground-truth labels for every statement. We establish a comprehensive benchmark evaluating state-of-the-art MLLMs, revealing a significant performance gap: even powerful models like GPT-4o struggle to distinguish truth from falsehood reliably. Our analysis of failure modes indicates that these models fail to ground language in visual social cues effectively and may be overly conservative in their alignment, highlighting the urgent need for novel approaches to building more perceptive and trustworthy AI systems.

\end{abstract}

%% file: sec/1_intro.tex
\section{Introduction}
\label{sec:intro}

Human society is built upon complex communication and interactions, with trust serving as the cornerstone for its efficient operation. However, deception, a pervasive social phenomenon~\cite{ekman2009telling}, profoundly impacts interpersonal relationships, economic activities, and even public safety. From everyday ``white lies'' to financial fraud and the spread of misinformation~\cite{shu2017fake}, accurately identifying deceptive intent in discourse is crucial for understanding the complexities of human society. It also presents a core challenge for building more intelligent and secure AI systems, such as advanced conversational agents and content moderation platforms \cite{picard1997affective}. To foster a future where AI and humans can collaborate seamlessly, equipping AI with sophisticated social perception and reasoning capabilities is indispensable.

\begin{figure}[t]
  \centering
   \includegraphics[width=0.99\linewidth]{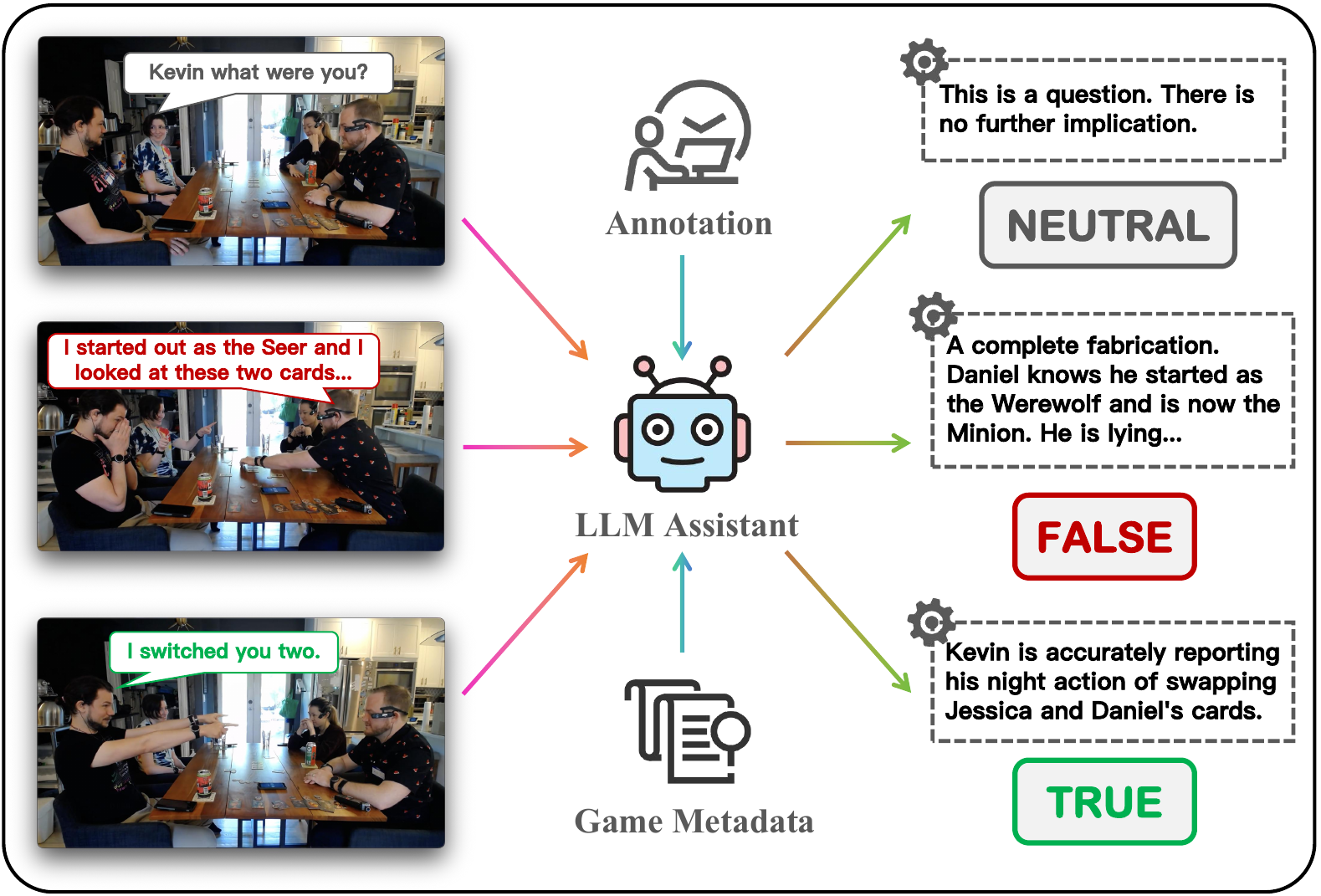}
   \caption{The MIVA Task Annotation Process. Starting with existing data and game metadata, we manually annotated the ``night actions." An automated, LLM-assisted pipeline then created a new multimodal MIVA dataset from the Werewolf game.}
    \label{fig:radar_chart}
     \vspace{-2ex}
\end{figure}

While deception detection has garnered research attention, existing work largely suffers from three major limitations. 
\textbf{(i) Lack of interactional context.} Prior studies often operate in isolation, analyzing single text snippets~\cite{ott2011finding}, unidirectional speech videos~\cite{perez2015deception, van2015deception}, or independent physiological signals~\cite{langleben2002brain, vrij2010pitfalls}. However, most real-world deception occurs within dynamic, interactive conversations. In such settings, deception is not a static, one-time act but a continuous process of real-time interaction with others' verbal and non-verbal feedback. 
\textbf{(ii) Simplification of social complexity. }
While prior work has made progress in two-person settings, such as the ``Box of Lies'' game~\cite{soldner2019box}, their interaction patterns are relatively structured (one person describes, another guesses). 
This fails to capture the complexity of authentic social deception, which frequently unfolds in messy, multi-party networks of coexisting alliances and rivalries. The social dynamics in these settings, including group pressure, covert collusion, and identity masking, introduce a level of complexity far exceeding simple question-answer paradigms~\cite{meta2022human}.
\textbf{(iii) Scarcity of verifiable ground truth.} A pivotal obstacle hindering the development of deception research is the lack of publicly available datasets with verifiable ground truth. In real-world scenarios, objectively annotating the precise moments of deception is often impossible, severely hampering the training and evaluation of predictive models~\cite{perez2015deception}. 

\begin{table*}[]
\centering
\scalebox{0.85}{
\begin{tabular}{l|l|cc|ccc}
\toprule  
\textbf{Dataset}                           & \textbf{Strategy Category} & \textbf{Utterances} & \textbf{Proportion} & \textbf{TRUE} & \textbf{FALSE} & \textbf{NEUTRAL} \\ \midrule  
\multirow{7}{*}{Ego4D} & Identity Declaration       & 53                  & 6.5\%               & 77.4\%        & 18.9\%         & 3.8\%            \\
                      & Evidence                   & 96                  & 11.7\%              & 57.3\%        & 11.5\%         & 31.3\%           \\
                      & Accusation                 & 89                  & 10.9\%              & 7.9\%         & 3.4\%          & 88.8\%           \\
                      & Interrogation              & 124                 & 15.1\%              & 4.8\%         & 0.8\%          & 94.4\%           \\
                      & Defense                    & 98                  & 12.0\%              & 28.6\%        & 11.2\%         & 60.2\%           \\
                      & Call for Action            & 52                  & 6.3\%               & 9.6\%         & 1.9\%          & 88.5\%           \\
                      & No Strategy                & 430                 & 52.5\%              & 2.6\%         & 0.9\%          & 96.5\%           \\ \midrule 
\multirow{7}{*}{Youtube}                   & Identity Declaration       & 42                  & 7.7\%               & 71.4\%        & 28.6\%         & 0.0\%            \\
                                           & Evidence                   & 60                  & 11.0\%              & 50.0\%        & 30.0\%         & 20.0\%           \\
                                           & Accusation                 & 104                 & 19.2\%              & 19.2\%        & 12.5\%         & 68.3\%           \\
                                           & Interrogation              & 76                  & 14.0\%              & 3.9\%         & 2.6\%          & 93.4\%           \\
                                           & Defense                    & 69                  & 12.7\%              & 36.2\%        & 13.0\%         & 50.7\%           \\
                                           & Call for Action            & 39                  & 7.2\%               & 2.6\%         & 5.1\%          & 92.3\%           \\
                                           & No Strategy                & 220                 & 40.5\%              & 7.3\%         & 2.7\%          & 90.0\% \\ \bottomrule          
\end{tabular}
}
\caption{Distribution of veracity labels across persuasive strategy categories in two datasets.}
\label{tab:ego4d_stats}
\vspace{-2ex}
\end{table*}

To address these limitations, we introduce the social deduction game \textbf{Werewolf} as our experimental paradigm. This complex multi-party game serves as a controlled yet ecologically valid environment that elicits natural, high-stakes deception, while crucially providing objective, deterministic ground truth derived from the game's rules and outcomes.
Building upon this paradigm and preview work~\cite{lai2023werewolf}, we construct our dataset using a novel semi-automated pipeline to achieve fine-grained annotation. To mitigate the uncertainty that comes with directly inferring player roles, we first manually annotated the crucial \textbf{``night actions"} for each game. We then employed an LLM to efficiently parse game events and player statements against the known game state (e.g., player roles, voting records). This process generates initial veracity labels, which are then rigorously verified against the game's ground truth to ensure high-fidelity, objective annotations.

To evaluate modern MLLMs on this challenging task, we introduce novel multimodal Chain-of-Thought (CoT) prompting strategies~\cite{wei2022chain, zhang2023multimodal}. Specifically, our Face-Focused CoT and Body-Focused CoT prompts instruct the model to first perform explicit visual analysis, interpreting non-verbal cues like eye contact, gestures, and posture, before reasoning toward a final veracity judgment.

Our analysis reveals profound limitations in the social intelligence of current MLLMs, which we categorize into three core deficiencies. First, they exhibit an overly conservative alignment, preventing them from making decisive, high-stakes judgments about veracity. 
Second, they lack a functional ``Theory of Mind"~\cite{kosinski2023theory}, the ability to infer others' hidden beliefs and strategic intentions, meaning they are unable to create and maintain a mental model of what other players know, believe, or want to achieve. 
Third, they struggle to distinguish salient social signals from distracting noise, often misreading subtle visual cues or over-relying on literal linguistic content. In short, current MLLMs act as powerful knowledge engines but are not yet competent social agents, failing to grasp the dynamic, intention-driven world behind the words.
In summary, our contributions are as follows:

\begin{itemize}
    \item We introduce and formalize the \textbf{Multimodal Interactive Veracity Assessment (MIVA)} task. To support this, we manually annotate \textbf{``night actions"} and construct a new multimodal dataset based on the Werewolf game, featuring synchronized video and text with verifiable ground-truth annotations created via an automatic, LLM-assisted pipeline.

    \item We perform a comprehensive benchmark of state-of-the-art MLLMs on this task, uncovering their substantial limitations. We then conduct an in-depth analysis of the models' failure modes and present preliminary explorations of potential approaches to enhance their performance on this challenging task.
    
\end{itemize}



%% file: sec/2_Related_Work.tex
\section{Related Work}
\label{sec:related_work}

\subsection{Multimodal Social Interaction}

The field of multimodal social interaction aims to endow machines with the ability to understand nuanced human behavior by integrating signals from various modalities, including language, acoustics, and vision. Seminal work in this area has focused on analyzing observable social phenomena in group settings. For instance, the AMI Meeting Corpus~\cite{carletta2005ami} enabled research into summarizing multi-party conversations and analyzing group dynamics like dominance and engagement~\cite{pentland2010honest}. More recent large-scale datasets, such as CMU-MOSI~\cite{zadeh2016mosi} and CMU-MOSEI~\cite{zadeh2018multi}, have pushed the boundaries of modeling subjective states like sentiment and emotion from video monologues. Complementing these efforts, recent social analysis work built upon the Werewolf Game has also emerged, with separate studies focusing on persuasion behaviors~\cite{lai2023werewolf}, social interactions~\cite{lee2024modeling}, gesture understanding~\cite{cao2025socialgesture}, and online interaction understanding~\cite{li2025towards}. 
These works have established powerful baselines for multimodal fusion and representation learning. 

However, prior research has predominantly focused on readily observable behaviors or affective states. Our work diverges by targeting a latent, cognitive, and intentional state: \textit{veracity}. Discerning truth from deception requires reasoning beyond surface-level emotions or engagement cues, demanding a deeper understanding of a speaker's knowledge, intentions, and strategic motivations within a high-stakes, adversarial context.

\subsection{Deception Detection}

Computational deception detection has a rich history, with early research often focusing on single modalities. Text-based approaches have successfully identified deceptive language in domains like online reviews~\cite{ott2011finding, li2014towards} and legal documents~\cite{fornaciari2013automatic}. Other lines of research have explored vocal cues like pitch and speech rate~\cite{montacie2016prosodic, litvinova2017building}, visual cues such as facial micro-expressions~\cite{ekman2009telling, yan2014casme} and eye movements~\cite{perez2015verbal}; and physiological indicators like fMRI data~\cite{langleben2002brain}.

Recognizing the multifaceted nature of deception, the field has increasingly shifted towards multimodal approaches. Datasets like the Real-Life Trial corpus~\cite{perez2015deception} provided a valuable resource with non-interactive, monologue-style videos. A significant step towards interactive settings was the  ``Box of Lies" dataset~\cite{zadeh2019boxoflies}, which introduced a two-person game to elicit deceptive behavior in a conversational context.

Our MIVA benchmark builds upon these foundations but pushes the frontier in two critical dimensions. First, we move from non-interactive monologues and structured dyadic conversations to the far more complex and ecologically valid domain of messy, multi-party social interactions, where alliances and rivalries coexist. Second, by leveraging a social deduction game, we provide a large-scale dataset with objective, verifiable ground truth for every statement, addressing a long-standing challenge of annotation ambiguity in deception research.

\subsection{Computational Modeling of Deduction Games}

Using complex games as crucibles for advancing AI has a long and successful history, from perfect-information games like Chess and Go~\cite{silver2016mastering} to imperfect-information games like Poker~\cite{brown2018superhuman}. Recently, the focus has shifted towards games requiring sophisticated communication, negotiation, and social reasoning. The most prominent example is the game of Diplomacy, where models like Cicero have demonstrated human-level performance by integrating strategic reasoning with natural language processing~\cite{meta2022human}. Similarly, prior work has explored agent-based strategies for playing text-based versions of the Werewolf game, focusing on game theory and logical deduction from dialogue~\cite{toriumi2016ai}.

While this body of work is impressive, 
our work offers a fundamentally different perspective. We are not building an agent to `play' the game; instead, we are using the game as a challenging benchmark to evaluate the multimodal social perception of modern MLLMs. By providing synchronized video and text, we ask a novel question: \textit{Can these powerful models read the room by grounding linguistic content in the visual signals of a complex social environment?} To our knowledge, MIVA is the first benchmark to bridge the gap between computational modeling of social deduction games and multimodal veracity assessment in complex, multi-party settings.

%% file: sec/3_Task_Dataset.tex
\begin{figure}[h!]
    \centering
    \fbox{
    \begin{minipage}{0.95\columnwidth}
    \small
        \textbf{Game Rules:} A detailed description of all roles, objectives, and the sequence of night actions in ``One Night Ultimate Werewolf.
    \vspace{0.5em}
    \hrule
    \vspace{0.5em}
  \textbf{Assigned Role:} Expert analyst of ``One Night Ultimate Werewolf" with a deep understanding of game theory and human deception.
    \vspace{0.5em}
    \hrule
    \vspace{0.5em}
    \textbf{Inputs:}
    \begin{itemize}
        \item \textbf{Game State (JSON):} Complete player list, start/end roles, voting outcomes, and \textbf{night actions (manually annotated)}.
        \item \textbf{Dialogue Log (CSV):} Full game transcript with timestamps and speaker identities.
    \end{itemize}
    \vspace{0.5em}
    \hrule
    \vspace{0.5em}
    \textbf{Label Definitions:}
    \begin{itemize}
        \item \textbf{\texttt{TRUE}:} The statement aligns with the speaker's known reality. (\textit{e.g., A real Villager claiming to be a Villager.})
        \item \textbf{\texttt{FALSE}:} The statement contradicts the speaker's known reality; an active lie. (\textit{e.g., A Werewolf claiming to be a Seer.})
        \item \textbf{\texttt{NEUTRAL}:} Truthfulness cannot be determined from the speaker's knowledge. Includes questions, opinions, and guesses. (\textit{e.g., "I think Paul is suspicious."})
    \end{itemize}
    \vspace{0.5em}
    \hrule
    \vspace{0.5em}
    \textbf{Core Principle:} Judgment must be strictly based on the private information the speaker possessed at the exact moment of the statement. 
    \vspace{0.5em}
    \hrule
    \vspace{0.5em}
    \textbf{Workflow:}
    \begin{enumerate}
        \item \textbf{Global Analysis:} Reconstruct the full game sequence and each player's knowledge based on the Game State data.
        \item \textbf{Line-by-Line Judgment:} For each utterance, determine its veracity based on the speaker's known facts at that moment.
        \item \textbf{Label Assignment:} Assign a label from \{\texttt{TRUE, FALSE, NEUTRAL}\} and provide a concise justification.
    \end{enumerate}
    \vspace{0.5em}
    \hrule
    \vspace{0.5em}
    \textbf{Required Output Columns:}
    \texttt{Timestamp, Speaker, Line, Truthfulness, Explanation}
    \end{minipage}
    }
    \caption{Overview of the semi-automated annotation prompt. The LLM is tasked to act as an expert analyst, following a strict workflow to produce verifiable veracity labels and explanations. The full prompt is in Appendix A.1.}
    \label{fig:annotation_rules}
\end{figure}

\section{The MIVA Benchmark}
\label{sec:dataset}

To facilitate the study of veracity assessment in multi-party interactions, we construct the \textbf{MIVA (Multimodal Interactive Veracity Assessment)} benchmark. We extend two existing social deduction game datasets with fine-grained, verifiable veracity annotations.

\subsection{Dataset Curation and Annotation}

\paragraph{Data Sources.}
Our benchmark is built upon two corpora featuring the game ``One Night Ultimate Werewolf'':
\begin{itemize}
    \item \textbf{Ego4D-MIVA}: We use a subset of the Ego4D social dataset~\cite{grauman2022ego4d}, consisting of 40 game sessions recorded from a third-person perspective to ensure all participants are visible. Our annotated test set comprises 819 utterances from a split defined in prior work~\cite{lai2023werewolf}.
    \item \textbf{YouTube-MIVA}: This corpus is curated from 151 game videos collected from YouTube~\cite{lai2023werewolf}, from which a subset of 543 utterances was selected for our detailed veracity annotation.
\end{itemize}
Both datasets provide synchronized videos, transcripts, and essential game metadata such as player roles and voting outcomes.

\paragraph{Annotation Pipeline.}
To address the inherent ambiguity of deducing night events from start and end roles alone, we introduce the manual annotation of \textbf{``night actions''}, which records the precise target of every action taken during the night phase (e.g., which player the Robber targeted).

We then introduce a novel semi-automated pipeline, employing a state-of-the-art LLM, Gemini-2.5-Pro, to parse game events and transcripts against the deterministic ground truth of the game (e.g., roles, night actions, voting results) and generate objective labels. 


\paragraph{Annotation Principles.}
The cornerstone of our annotation is that \textbf{veracity is judged based on the speaker's private knowledge state at the moment of utterance}. To operationalize this, we developed a detailed \textbf{rulebook} (see Appendix A) that guides the labeling process. The framework is summarized in Figure~\ref{fig:annotation_rules}.
For quality assurance, we also conduct a preliminary human validation study on a subset (5\%) of the data, achieving a human-LLM agreement (87.8\% accuracy).


\begin{figure}[]
    \centering
    \fbox{
    \begin{minipage}{0.95\columnwidth}
    \small
    \vspace{0.5em}
    \textbf{Context Provided to Model:}
    \begin{enumerate}
        \item \textbf{Game Rules:} A detailed description of all roles, objectives, and the sequence of night actions in ``One Night Ultimate Werewolf."
        \item \textbf{Conversation History:} The transcript of the discussion leading up to the current moment.
        \item \textbf{Current Utterance and Picture:} The specific statement to be analyzed, including the speaker and timestamp.
    \end{enumerate}
    \vspace{0.5em}
    \hrule
    \vspace{0.5em}
    \textbf{Hierarchical Task Definition:}
    For the current utterance, the model must perform two sequential labeling tasks:
    \begin{itemize}
        \item \textbf{Label 1 (Strategy):} Classify the statement across six persuasive strategies.
        \item \textbf{Label 2 (Veracity):} Determine the final veracity from \{\texttt{TRUE, FALSE, NEUTRAL}\}.
    \end{itemize}
    \vspace{0.5em}
    \hrule
    \vspace{0.5em}
    \textbf{Required Output Format (JSON):}
    
    \par\noindent 
    \texttt{\{} \\
    \texttt{\ \ "identity\_declaration": "TRUE or FALSE",} \\
    \texttt{\ \ "accusation": "TRUE or FALSE",} \\
    \texttt{\ \ "defense": "TRUE or FALSE",} \\
    \texttt{\ \ "evidence": "TRUE or FALSE",} \\
    \texttt{\ \ "interrogation": "TRUE or FALSE",} \\
    \texttt{\ \ "call\_for\_action": "TRUE or FALSE",} \\
    \texttt{\ \ "veracity": "Choose one: TRUE, FALSE, or NEUTRAL.",} \\
    \texttt{\ \ "reasoning": "Provide a brief justification..."} \\
    \texttt{\}}

    \end{minipage}
    }
    \caption{Summary of the MLLM evaluation prompt. The model receives comprehensive context and is tasked with a hierarchical analysis of both persuasive strategy and veracity, with a required structured JSON output.}
    \label{fig:evaluation_prompt_summary}
\end{figure}

\subsection{Task Definition and Evaluation}
We formally define the MIVA task as follows: Given a multimodal clip of a player's utterance, including the conversational history and the game rules, the objective is to predict its veracity label from the set $\{\texttt{TRUE, FALSE, NEUTRAL}\}$.

For model evaluation, we adopt a hierarchical reasoning approach. The MLLM is first prompted to identify the presence of six persuasive strategies (\textit{Identity Declaration}, \textit{Accusation}, \textit{Defense}, \textit{Evidence}, \textit{Interrogation}, \textit{Call for Action}) defined in prior work~\cite{lai2023werewolf}. Subsequently, it must determine the final veracity label. This approach encourages the model to first reason about the statement's strategic function before assessing its truthfulness. The model is required to output its analysis in a structured JSON format. We provide the prompt template in Figure~\ref{fig:evaluation_prompt_summary}. The detailed evaluation prompt is provided in the Appendix A.2.

\begin{table*}[!ht]
\centering
\caption{Evaluation results for the Persuasive Strategy Classification task on the Ego4D and YouTube datasets. We report F1 scores for each of the six categories, along with the average F1 score and Joint Accuracy (all six labels predicted correctly). Best performance in each row is in \textbf{bold}.}
\label{tab:strategy_results}
\resizebox{\textwidth}{!}{%
\begin{tabular}{l|l|cccccc}
\toprule
\textbf{Dataset} & \textbf{Metric} & \textbf{GPT-4o-mini} & \textbf{GPT-4o} & \textbf{GPT5-nano} & \textbf{Gemini-2.5-pro} & \textbf{Deepseek-v3} & \textbf{Claude-3.5-haiku} \\
\midrule
\multirow{8}{*}{Ego4D}   & Identity Declaration & \textbf{84.4}                   & 67.5                       & 74.6                          & 68.0                               & 82.4                            & 71.6                                 \\
                         & Accusation           & 45.1                            & \textbf{47.6}              & 40.0                          & 45.6                               & 47.6                            & 44.4                                 \\
                         & Defense              & 22.2                            & 29.9                       & \textbf{36.5}                 & 34.3                               & 29.0                            & 36.0                                 \\
                         & Evidence             & 49.0                            & 61.7                       & 60.7                          & \textbf{66.3}                      & 58.2                            & 50.6                                 \\
                         & Interrogation        & 72.0                            & 82.3                       & 81.0                          & \textbf{84.1}                      & 83.5                            & 76.5                                 \\
                         & Call for Action      & 51.0                            & \textbf{65.0}              & 50.7                          & 52.6                               & 57.1                            & 45.4                                 \\ \cmidrule{2-8}
                         & Avg F1 Score         & 53.9                            & 59.0                       & 57.3                          & 58.5                               & \textbf{59.6}                   & 54.1                                 \\
                         & Joint Accuracy       & 62.9                            & \textbf{63.2}              & 58.4                          & 54.0                               & 62.8                            & 46.6                                 \\ \midrule
\multirow{8}{*}{Youtube} & Identity Declaration & \textbf{79.1}                   & 65.1                       & 76.2                          & 71.2                               & 78.1                            & 71.3                                 \\
                         & Accusation           & 59.1                            & 57.8                       & 57.0                          & \textbf{59.8}                      & 59.3                            & 58.6                                 \\
                         & Defense              & 34.0                            & 32.7                       & 45.5                          & \textbf{51.1}                      & 42.7                            & 49.7                                 \\
                         & Evidence             & 50.8                            & 56.2                       & 58.2                          & 54.8                               & \textbf{60.3}                   & 51.2                                 \\
                         & Interrogation        & 67.4                            & 73.8                       & \textbf{80.5}                 & 78.9                               & 72.5                            & 70.7                                 \\
                         & Call for Action      & 62.6                            & 66.0                       & 68.0                          & 65.5                               & \textbf{70.2}                   & 56.6                                 \\
                         \cmidrule{2-8}
                         & Avg F1 Score         & 58.8                            & 58.6                       & \textbf{64.2}                 & 63.5                               & 63.9                            & 59.7                                 \\
                         & Joint Accuracy       & 55.1                            & 52.5                       & 55.2                          & 49.0                               & \textbf{56.0}                   & 46.8   \\      
\bottomrule
\end{tabular}
}
\vspace{-2ex}
\end{table*}




\subsection{Dataset Analysis}
\label{sec:dataset_analysis}

We provide a statistical analysis of our annotated Dataset as an example in Table~\ref{tab:ego4d_stats}. The analysis reveals key characteristics with significant implications for model evaluation.

\vspace{-1ex}

\paragraph{Data Imbalance and Concentration of Verifiable Claims.}
A primary characteristic of our dataset is the sparsity of factual claims. The discourse is composed of \texttt{NEUTRAL} utterances, such as questions, opinions, and strategic probes, rather than directly verifiable statements. For instance, in categories like \textit{Interrogation} and \textit{No Strategy}, over 90\% of statements are \texttt{NEUTRAL}. 
While sparse overall, verifiable claims (\texttt{TRUE}/\texttt{FALSE}) are densely concentrated in specific, high-stakes categories. Statements in \textit{Identity Declaration} and \textit{Evidence} are predominantly factual assertions, making them the primary battlegrounds for veracity assessment. 
Consequently, a fundamental challenge for any AI model is to first distinguish the fact-checkable ``signals" (\texttt{TRUE}/\texttt{FALSE}) from the predominant ``noise" of neutral conversation before veracity assessment can be made.


These distributional properties underscore the inadequacy of simple accuracy as an evaluation metric and necessitate the use of metrics like \textit{F1-score / macro F1-score} and a specific focus on \textit{binary accuracy} for \texttt{TRUE}/\texttt{FALSE} instances to meaningfully evaluate a model's ability to handle the sparse but critical moments of deception.

\begin{table*}[t]
\centering
\caption{Prediction results for the MIVA task on the Ego4D and YouTube datasets. We report Accuracy, Accuracy on only `TRUE'/`FALSE' samples (Binary), and macro-averaged Precision, Recall, and F1-score. Best performance is in \textbf{bold}.}
\label{tab:miva_results}
\resizebox{\textwidth}{!}{%
\begin{tabular}{l|l|cccccc}
\toprule
\textbf{Dataset} & \textbf{Metric} & \textbf{GPT-4o-mini} & \textbf{GPT-4o} & \textbf{GPT5-nano} & \textbf{Gemini-2.5-pro} & \textbf{Deepseek-v3} & \textbf{Claude-3.5-haiku} \\
\midrule
\multirow{5}{*}{Ego4D}   & Accuracy          & 66.3                            & \textbf{74.0}              & 67.9                          & 68.6                               & 71.5                            & 71.7                                 \\
                         & Accuracy (Binary) & \textbf{39.4}                   & 27.6                       & 14.2                          & 26.0                               & 7.9                             & 4.7                                  \\  \cmidrule{2-8}
                         & Macro-Precision   & 43.6                            & \textbf{52.7}              & 39.1                          & 43.2                               & 41.5                            & 42.0                                 \\
                         & Macro-Recall      & 47.3                            & \textbf{52.5}              & 37.9                          & 46.3                               & 35.8                            & 35.6                                 \\
                         & Macro-F1          & 44.5                            & \textbf{51.2}              & 38.1                          & 44.2                               & 35.7                            & 35.2                                 \\ \midrule
\multirow{5}{*}{Youtube} & Accuracy          & 62.5                            & 64.4                       & 60.7                          & \textbf{65.6}                      & 62.9                            & 61.9                                 \\
                         & Accuracy (Binary) & \textbf{37.4}                   & 15.0                       & 14.3                          & 27.9                               & 5.4                             & 2.7                                  \\  \cmidrule{2-8}
                         & Macro-Precision   & 50.7                            & 48.1                       & 42.7                          & 54.6                               & \textbf{59.3}                   & 49.2                                 \\
                         & Macro-Recall      & \textbf{50.4}                   & 41.7                       & 38.5                          & 49.6                               & 37.3                            & 35.3                                 \\
                         & Macro-F1          & 50.5                            & 42.1                       & 38.6                          & \textbf{50.7}                      & 35.6                            & 32.1   \\
\bottomrule
\end{tabular}
}
\end{table*}

\paragraph{Influence of Player Expertise.}
Interestingly, we observed a distinction between the player demographics. Players in the Ego4D dataset tended to be novices, whereas the YouTube data featured more experienced players.
The novice-expert divide between the corpora reveals clear differences in strategic gameplay. Expert players (YouTube) exhibit higher strategic density, with a lower proportion of non-strategic utterances than novices (40.5\% vs. 52.5\%). More critically, experts engage in more sophisticated deception. The rate of false \textit{Evidence} claims among experts is nearly three times that of novices (30.0\% vs. 11.5\%), and they lie more frequently when declaring their identity (28.6\% vs. 18.9\%). 
This validates that our MIVA benchmark spans a meaningful spectrum of difficulty, with the YouTube corpus representing a more challenging testbed for deception detection.

%% file: sec/4_Benchmark.tex
\section{Experiments}
\label{sec:experiments}

In this section, we present a comprehensive evaluation of state-of-the-art Multimodal Large Language Models (MLLMs) on our proposed MIVA benchmark. We first establish baseline performance on our two core tasks, then conduct a series of ablation studies to analyze the effects of visual and temporal information.

\subsection{Experimental Setup}
\label{sec:models}

\paragraph{Models.}
Our evaluation focuses on leading proprietary, closed-source MLLMs, as they currently represent the state-of-the-art in multimodal understanding and reasoning capabilities. We select a diverse range of powerful models, including GPT-4o~\cite{openai2024gpt4o}, GPT-4o-mini~\cite{openai2024gpt4omini}, Gemini-2.5-pro~\cite{Google2025Gemini2_5}, and Claude-3.5-haiku~\cite{Anthropic2024Claude3_5}, along with Deepseek-v3~\cite{liu2024deepseek} and GPT5-nano~\cite{openai2025gpt5nano}.
All experiments were conducted using the models' official APIs between July and August 2025.

\paragraph{Evaluation Metrics.}
Given the class imbalance observed in the dataset (Sec.~\ref{sec:dataset_analysis}), following previous work~\cite{lai2023werewolf}, we use the \textit{F1-score / macro-averaged F1-score} as our primary metric for both tasks to ensure that performance on minority classes is adequately represented. For the strategy classification task, we also report \textit{Joint Accuracy}, the percentage of utterances where all six strategy labels are correctly predicted. For the MIVA task, we report overall \textit{Accuracy} and \textit{Binary Accuracy} (correctly classifying only `TRUE' or `FALSE' instances) to specifically measure the model's ability to handle high-stakes judgments.

\subsection{Main Benchmark Results}
\label{sec:benchmark_results}

We evaluate the selected models on two hierarchical tasks: first, identifying persuasive strategies, and second, the core MIVA task of veracity assessment.

\paragraph{Performance on Persuasive Strategy.}
The classification of persuasive strategies proved to be a challenging task with no single dominant model across all conditions (Table~\ref{tab:strategy_results}). On the Ego4D dataset, performance among the top models was tightly contested, with Deepseek-v3 achieving the highest average F1 score (59.6), while GPT-4o attained the best Joint Accuracy (63.2\%), indicating better precision. A clearer hierarchy emerged on the YouTube dataset, where GPT5-nano established itself as the leader with a significantly higher average F1 score of 64.2. Across both datasets, models consistently performed well on categories with distinct linguistic patterns (\textit{e.g.}, \textit{Interrogation}), yet universally struggled with the nuanced language of \textit{Defense}, marking it as a key area for future work.

\paragraph{Performance on MIVA.}
For our core veracity assessment task, GPT-4o demonstrated the best overall performance on the Ego4D dataset, achieving the highest Macro-F1 score (51.2), suggesting better holistic reasoning (Table~\ref{tab:miva_results}). However, a deeper analysis of the Accuracy (Binary) metric, which isolates the critical \texttt{TRUE}/\texttt{FALSE} judgments, reveals a crucial distinction. GPT-4o-mini, despite its lower overall score, significantly outperforms all competitors in this specific sub-task (39.4\% on Ego4D).

This discrepancy exposes a shared pathology in most large MLLMs: First, an overly conservative alignment that causes them to evade high-stakes decisions by defaulting to the ``safe" \texttt{NEUTRAL} class, a behavior confirmed by their low Macro-Recall scores. Thus, while GPT-4o exhibits the best holistic comprehension of the discourse, GPT-4o-mini is uniquely adept at the pivotal skill of discriminating truth from deception.

Second, they lack a functional ``Theory of Mind"~\cite{kosinski2023theory}, the ability to infer others' hidden beliefs and strategic intentions. This means they cannot maintain a mental model of what other players know or want to achieve. Consequently, while the models perform well on straightforward, non-factual statements that can be heuristically classified as \texttt{NEUTRAL}, their performance collapses when faced with high-stakes factual claims that require a deep, inferential understanding of the speaker's knowledge state. 

In summary, it is crucial to note that the best-performing model, GPT-4o, only achieved a Macro-F1 score of 51.2 on Ego4D. This score, while leading the pack, is far from reliable and underscores the profound challenge. It clearly indicates that the social reasoning capabilities of even the most advanced MLLMs are still in their infancy and require substantial improvement.


\begin{table*}[t]
\centering
\caption{The impact of different visual input strategies on GPT-4o-mini's performance for the MIVA task. ``Vision" refers to the full frame, while ``Face-CoT" and ``Body-CoT" use our CoT prompting.}
\label{tab:modality_effects}
\resizebox{0.6\textwidth}{!}{%
\begin{tabular}{l|l|cccc}
\toprule
\textbf{Dataset} & \textbf{Metric} & \textbf{Text-only} & \textbf{Vision} & \textbf{Face-CoT} & \textbf{Body-CoT} \\
\midrule
\multirow{5}{*}{Ego4D}   & Accuracy          & 66.3                          & 69.4                       & \textbf{74.3}            & 71.5                     \\
                         & Accuracy (Binary) & \textbf{39.4}                 & 38.6                       & 32.3                     & 37.8                     \\ \cmidrule{2-6}
                         & Macro-Precision   & 43.6                          & 46.1                       & \textbf{47.9}            & 46.5                     \\
                         & Macro-Recall      & 47.3                          & \textbf{48.9}              & 46.4                     & 48.3                     \\
                         & Macro-F1          & 44.5                          & 47.0                       & 47.1                     & \textbf{47.2}            \\ \midrule
\multirow{5}{*}{Youtube} & Accuracy          & 62.5                          & 63.2                       & \textbf{67.2}            & 64.7                     \\
                         & Accuracy (Binary) & \textbf{37.4}                 & 29.3                       & 27.9                     & 31.3                     \\ \cmidrule{2-6}
                         & Macro-Precision   & 50.7                          & 49.2                       & \textbf{53.3}            & 51.8                     \\
                         & Macro-Recall      & \textbf{50.4}                 & 47.1                       & 48.1                     & 48.9                     \\
                         & Macro-F1          & \textbf{50.5}                 & 47.9                       & 49.7                     & 50.0                     \\
\bottomrule
\end{tabular}
}
\end{table*}

\begin{table*}[t]
\centering
\caption{Results of temporal ablation studies on the Ego4D-MIVA dataset using GPT-4o-mini. ``w/o History-info" removes the text history. ``Vision-1-Frame" is the default multimodal setting, compared with ``Vision-3-frame".}
\label{tab:temporal_effects}
\resizebox{\textwidth}{!}{%
\begin{tabular}{l|l|cc||cccc}
\toprule
\textbf{Task} & \textbf{Metric} & \textbf{Default w/ History} & \textbf{w/o History-info} & \textbf{Default (1-Frame)} & \textbf{w/ Vision (1-Frame)} & \textbf{w/ Vision (3-frame)} \\
\midrule
\multirow{8}{*}{Strategy} & Identity Declaration & \textbf{84.4}                   & 81.1             & \textbf{84.4}                   & 77.5           & 73.2           \\
                          & Accusation           & 45.1                            & \textbf{47.3}    & 45.1                            & \textbf{49.4}  & 46.6           \\
                          & Defense              & 22.2                            & \textbf{22.7}    & 22.2                            & \textbf{23.3}  & 18.2           \\
                          & Evidence             & 49.0                            & \textbf{56.1}    & \textbf{49.0}                   & 46.3           & 44.4           \\
                          & Interrogation        & 72.0                            & \textbf{77.6}    & \textbf{72.0}                   & 63.2           & 63.8           \\
                          & Call for Action      & 51.0                            & \textbf{53.1}    & \textbf{51.0}                   & 46.8           & 50.3           \\ \cmidrule{2-7}
                          & Avg F1 Score         & 53.9                            & \textbf{56.3}    & \textbf{53.9}                   & 51.1           & 49.4           \\
                          & Joint Accuracy       & \textbf{62.9}                   & 61.9             & \textbf{62.9}                   & 55.1           & 55.3          \\
\midrule \midrule
\multirow{5}{*}{MIVA}     & Accuracy             & 66.3                            & \textbf{72.2}    & 66.3                            & \textbf{69.4}  & 68.9           \\
                          & Accuracy (Binary)    & \textbf{39.4}                   & 13.4             & \textbf{39.4}                   & 38.6           & 35.4           \\ \cmidrule{2-7}
                          & Macro-Precision      & 43.6                            & \textbf{46.4}    & 43.6                            & \textbf{46.1}  & 45.6           \\
                          & Macro-Recall         & \textbf{47.3}                   & 42.3             & 47.3                            & \textbf{48.9}  & 48.7           \\
                          & Macro-F1             & \textbf{44.5}                   & 42.1             & 44.5                            & \textbf{47.0}  & 46.6           \\
\bottomrule
\end{tabular}
}
\end{table*}

\begin{figure*}[h]
  \centering
   \includegraphics[width=0.85\linewidth]{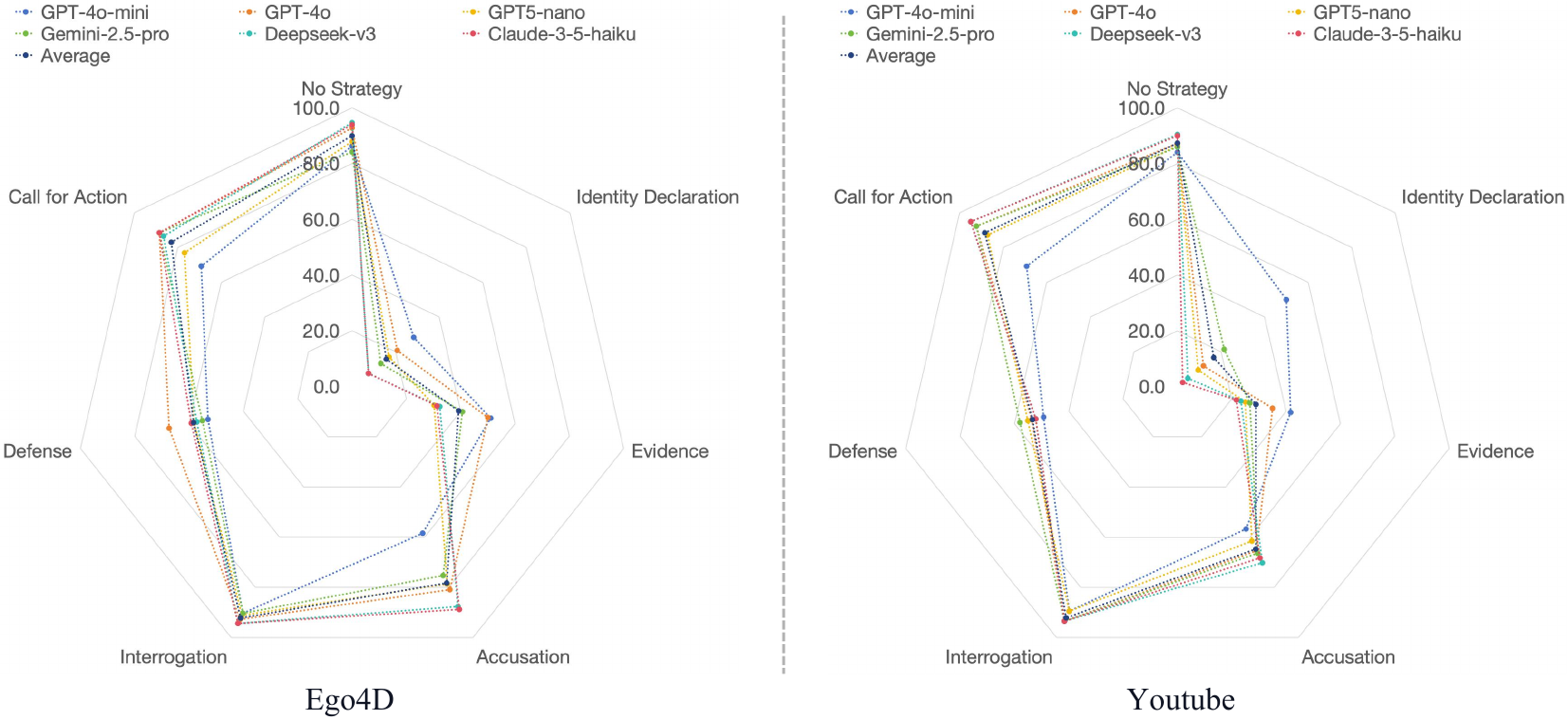}
   \caption{Radar Chart of Models’ Accuracy in the MIVA task across persuasive strategy categories in two datasets.}
    \label{fig:radar_chart}
\end{figure*}



\subsection{Effect of Visual Modality}
\label{sec:ablation_visual}

To understand how MLLMs utilize visual information, we conducted an ablation study on GPT-4o-mini using four different input settings: (1) \textbf{Text-only}: The model receives only the transcript. (2) \textbf{Vision}: The model receives the transcript and a single video frame from the utterance. (3) \textbf{Face-CoT}: The model is prompted to first analyze the speaker's facial expression in a Chain-of-Thought manner before making a prediction. (4) \textbf{Body-CoT}: The model is prompted to first analyze the speaker's body language and posture. The full CoT prompts are in Appendix A.3 / A.4.

Our results, presented in Table~\ref{tab:modality_effects}, reveal that visual information acts as a ``double-edged sword''. On one hand, incorporating visual cues, particularly when guided by our CoT prompts, consistently improves the model's holistic comprehension of the dialogue. On the Ego4D dataset, all multimodal approaches substantially outperform the text-only baseline in overall Macro-F1, with Body-CoT achieving the highest score of 47.2 (+2.7 points). The Face-CoT method yields the best overall Accuracy (74.3\%).

On the other hand, all forms of visual input significantly degraded the model's performance on the pivotal task of distinguishing truth from lies. The `Accuracy (Binary)' metric, which measures performance on only \texttt{TRUE}/\texttt{FALSE} instances, was highest for the `Text-only' model across both datasets. 
While the Face-CoT and Body-CoT prompts generated accurate, detailed descriptions of visual cues, this structured analysis failed to correctly interpret their social meaning. This suggests that while models can ``see" and describe non-verbal signals, they sometimes struggle to effectively ground their final veracity judgment in this visual evidence.

\subsection{Effect of Temporal Information}
\label{sec:ablation_temporal}

We investigate the role of temporal context by comparing GPT-4o-mini's performance under different conditions: (1) the default setting with full conversation history, (2) removing the text history (\textbf{w/o History-info}), and (3) providing multiple video frames (\textbf{1-frame vs. 3-frame}).

The results on Ego4D, shown in Table~\ref{tab:temporal_effects}, reveal a crucial dichotomy. For strategy classification, removing textual history has minimal negative impact and sometimes even improves the Avg F1 score (+2.4). This suggests that identifying a statement's strategic function is largely a local phenomenon. In sharp contrast, removing the history causes a catastrophic drop in veracity assessment performance (Accuracy (Binary) plummets from 39.4\% to 13.4\%). This demonstrates that assessing truthfulness is a global task that fundamentally relies on contextual reasoning over the entire conversation. 

Furthermore, simply adding more video frames (3-frame) did not provide additional benefits. In fact, for both the MIVA and strategy tasks, most metrics show a slight degradation with multi-frame input, reinforcing the conclusion that models struggle to distinguish social signals from distracting noise and effectively achieve robust visual grounding, even with more temporal visual data. 

\subsection{Performance Analysis across Categories}
\label{sec:analysis_categories}

As illustrated in Figure~\ref{fig:radar_chart}, all models exhibit a severe performance imbalance on the MIVA task across different strategy categories. They achieve high accuracy on ``easy" categories like \textit{No Strategy} and \textit{Interrogation}. However, this is largely misleading; since most statements in these categories are \texttt{NEUTRAL}, which does not reflect true reasoning ability.

Conversely, the models' true reasoning capabilities are tested in the ``hard," information-rich categories of \textit{Identity Declaration} and \textit{Evidence}, where performance plummets for all models. On the YouTube-MIVA dataset, for instance, the average accuracy on \textit{Identity Declaration} collapses to a mere 16.7\%. 
Within this challenging context, we observe a clear performance hierarchy: the GPT-4 family consistently holds an edge over competitors in these reasoning-intensive categories. GPT-4o-mini, in particular, shows a striking aptitude for identifying true and false identity claims from expert players (50.0\% accuracy), significantly outperforming even powerful models like Gemini-2.5-pro (21.4\%) and Deepseek-v3 (4.8\%) on this specific task. This underscores the need for future work to focus on the core logical reasoning required in these high-stakes contexts.


%% file: sec/5_Discussion_and_Conclusion.tex



\section{Discussion and Conclusion}
\label{sec:discussion_conclusion}

In this work, we introduced MIVA, a challenging benchmark for assessing veracity in multi-party social interactions. Through our novel dataset and comprehensive evaluation, we demonstrated that even state-of-the-art MLLMs have profound limitations in artificial social intelligence. Our analysis pinpoints three core deficiencies: (i) an overly conservative alignment that causes models to evade high-stakes judgments by defaulting to safe, neutral answers; (ii) a fundamental lack of a ``Theory of Mind" to infer the strategic intentions behind statements; and (iii) a critical inability to distinguish salient visual social cues from noise, leading to a failure of multimodal grounding. In short, current MLLMs function as powerful \textit{knowledge engines} but not yet as competent \textit{social agents}.

These findings highlight urgent and promising directions for future research. Bridging this gap requires moving beyond current paradigms to develop: context-adaptive alignment strategies that are less risk-averse in adversarial settings; new architectures with integrated Theory of Mind reasoning; and more robust methods for grounding language in non-verbal visual cues. Ultimately, addressing these challenges is a crucial step towards building the truly perceptive and trustworthy AI systems required for seamless human-AI collaboration.